\documentclass{article}
\usepackage{spconf,amsmath,amsfonts,graphicx,booktabs,array,hyperref,dblfloatfix,balance,enumitem}
\hypersetup{hidelinks}
\title{SAFESTYLE: CALIBRATED STYLE RESIDUAL INJECTION FOR CONTROLLABLE\\STYLE-LEAKAGE TRADE-OFF IN DIFFUSION STYLIZATION}

\name{Zhangping Yang$^{1}$\thanks{\textsuperscript{$\dagger$} Corresponding authors.},
    Min Li$^{1}$\textsuperscript{$\dagger$},
    Song Yan$^{2}$,
    Rong Gao$^{3}$,
    Xinliang Bi$^{1}$,
    Guanye Xiong$^{1}$,
    Yujie He$^{1}$}

\address{$^{1}$ Xi'an High-tech Research Institute, Shaanxi, China \\
$^{2}$ University of Science and Technology of China, Anhui, China \\
$^{3}$ Xi'an University of Architecture and Technology, Shaanxi, China \\
Emails: \{akame1027, proflimin, bxl970526, 13350433677, ksy5201314\}@163.com,\\
gary\_144@mail.ustc.edu.cn, eggymilk@163.com}

\begin{document}
\ninept
\emergencystretch=2em
\maketitle

\begin{abstract}
Reference-guided diffusion stylization aims to transfer visual style from a reference image while preserving the semantics specified by a text prompt. However, image conditioning often entangles transferable style cues with reference-specific content, leading to an inherent trade-off: stronger conditioning improves style fidelity but increases content leakage, whereas aggressive suppression reduces leakage at the cost of style expression. This challenge is further complicated by the distinct spatial organization of texture- and geometry-dominant styles. To address these issues, we propose SafeStyle, a training-free framework for calibrated style residual injection in frozen diffusion models. SafeStyle first estimates style-supported and content-associated subspaces from compact calibration sets, preserving their informative overlap while suppressing useless content variations. It then transports the purified style evidence over adaptive spatial granularity and constrains its effective influence through an explicit residual-norm budget. Experiments across texture- and geometry-dominant styles show that SafeStyle achieves a DINO style similarity of 0.432 while maintaining competitive text alignment. On a semantically disjoint leakage-stress benchmark, it further achieves a DINO style similarity of 0.474 with only 0.8\% semantic leakage, demonstrating an effective balance between style fidelity and reference-content suppression.
\end{abstract}

\begin{keywords}
Diffusion Models, Style Transfer, Reference Content Leakage, Training-free
\end{keywords}

\section{Introduction}
\begingroup
\setlength{\parskip}{0pt}

Reference guided diffusion stylization aims to generate images that follow a text prompt while inheriting the appearance of a reference image. Latent diffusion models~\cite{LDM,SDXL} and image conditioning mechanisms such as IP Adapter~\cite{IPAdapter} have made this paradigm increasingly practical. However, image conditions often entangle transferable style evidence, such as color, brushwork, and local patterns, with reference specific content, including subject identity, shape, and layout. Increasing the conditioning strength therefore improves style fidelity but also increases reference content leakage.

\begin{figure}[t]
    \centering
    \includegraphics[width=0.96\linewidth]{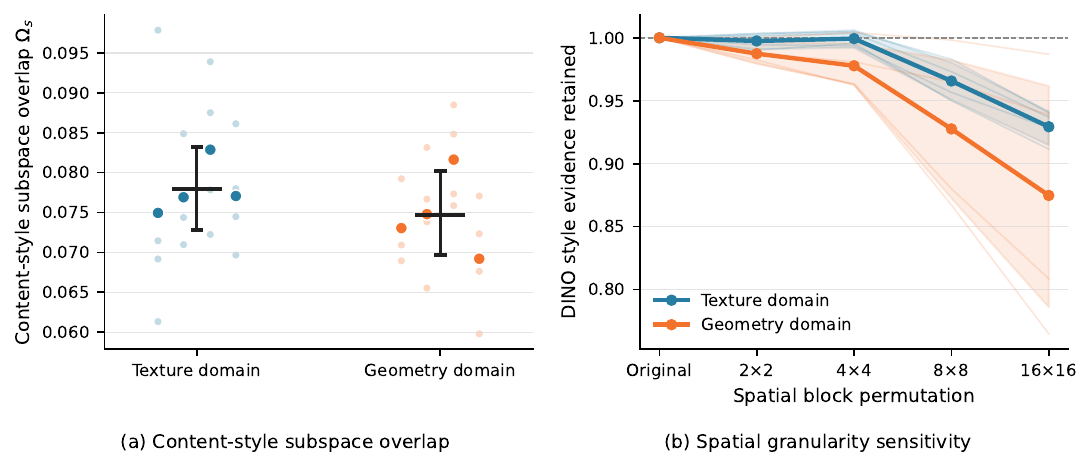}
    \vspace{-1.5mm}
    \caption{Style groups share non-zero content-style overlap but respond differently to spatial permutation, motivating common purification and adaptive transport granularity.}
    \label{fig:diagnosis}
\end{figure}

Existing methods address this problem from different perspectives. IP Adapter~\cite{IPAdapter} introduces decoupled image conditioning, while InstantStyle~\cite{InstantStyle} suppresses content associated features and selectively injects style information. Learned approaches such as DEADiff~\cite{DEADiff}, CSGO~\cite{CSGO}, StyleStudio~\cite{StyleStudio}, and StyleShot~\cite{StyleShot} improve style content disentanglement through specialized training, whereas training free methods such as StyleID~\cite{StyleID} and StyleAligned~\cite{StyleAligned} manipulate inversion or attention features without updating model parameters. Reference leakage is further reduced through feature masking, negative visual guidance, sampling control, or structure aware attention~\cite{MaskST,StyleKeeper,StyleSSP,OSASIS}. 
Recent work~\cite{imagdressing,imagpose} also emphasizes where style evidence is represented: StyleGallery~\cite{StyleGallery} aligns semantically clustered regions for arbitrary-reference transfer, while SEFS~\cite{SEFS} separates appearance from target geometry using low-resolution style crops and explicit structural conditions. CleanStyle~\cite{CleanStyle} is particularly related to our setting because it spectrally filters image conditions to suppress reference semantics. These advances establish the value of selective features and spatial support, but target-side constraints or aggressive filtering can limit text-only generation or discard transferable style evidence.

\begin{figure*}[t!]
    \centering
    \includegraphics[width=0.94\linewidth]{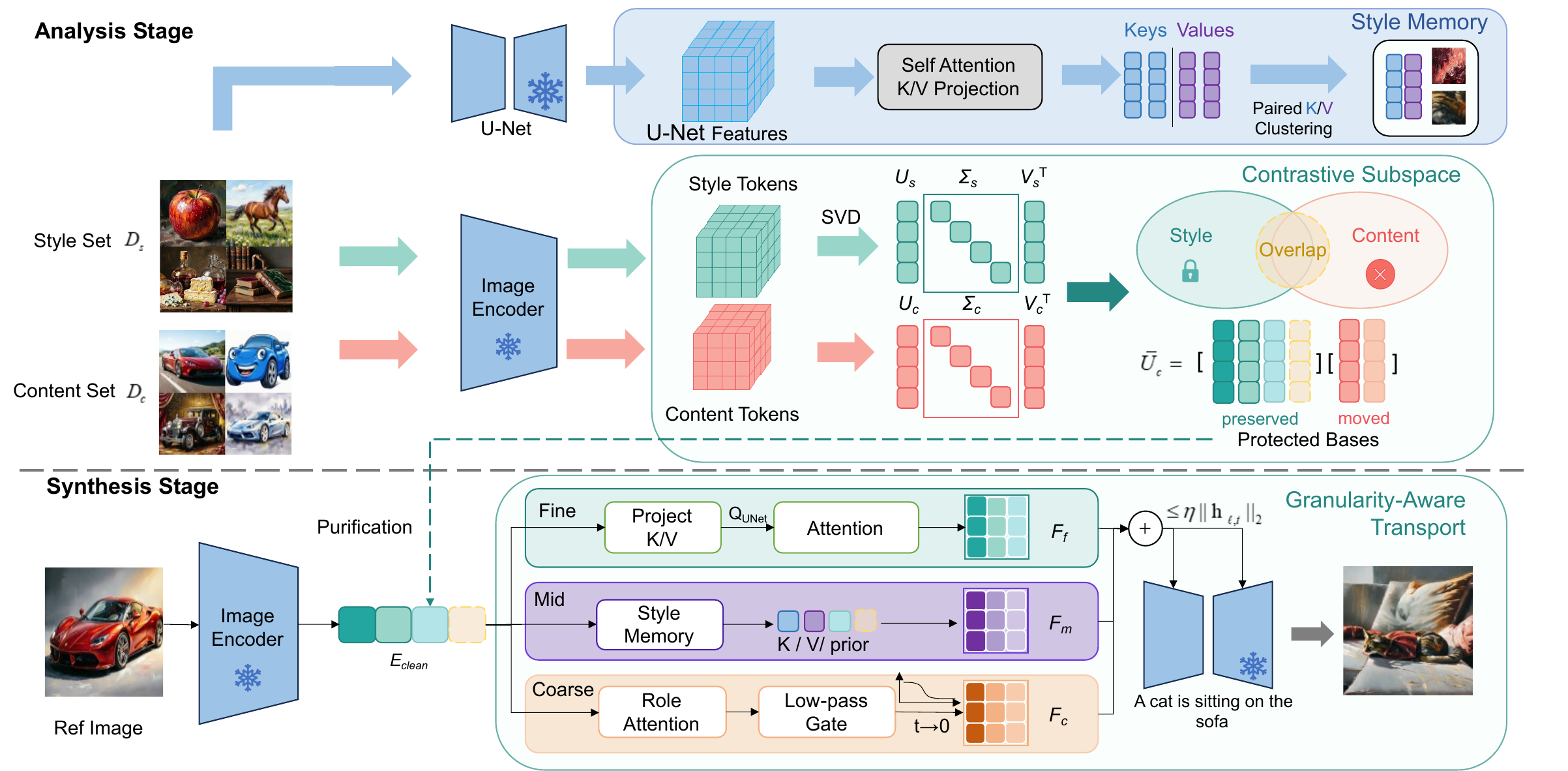}
    \vspace{-1.5mm}
    \caption{SafeStyle uses calibration sets to remove unsupported content evidence, transports the purified residual at adaptive spatial granularity, and injects it into a frozen U-Net under an actual residual-norm budget.}
    \label{fig:overview}
    \vspace{-2mm}
\end{figure*}

This limitation becomes more pronounced for styles with different spatial organizations. Texture dominant styles can often be represented by local appearance statistics, whereas geometry styles require evidence organized over larger spatial supports. As shown in Fig.~\ref{fig:diagnosis}, our diagnostics reveal two complementary properties. First, style and content associated subspaces retain a nonzero overlap across style groups, indicating that some content correlated directions also carry useful style evidence. Second, different styles exhibit distinct sensitivity to spatial permutation, suggesting that the surviving evidence should not be transferred at a fixed granularity. Therefore, effective reference stylization requires two coordinated decisions: determining which evidence should be retained and determining at what spatial support it should be delivered.

\begin{figure}[t]
    \centering
    \includegraphics[width=0.91\linewidth]{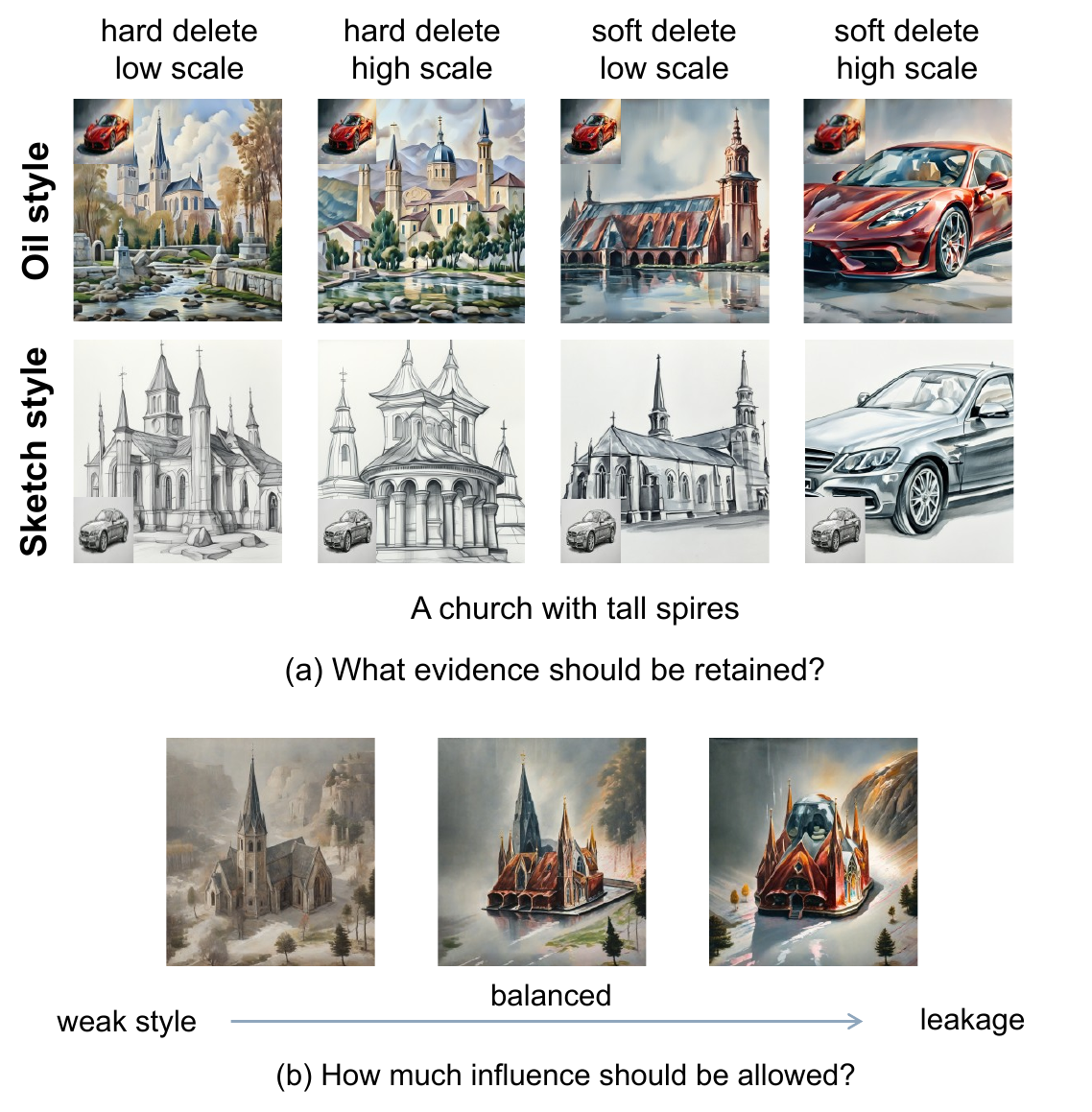}
    \vspace{-4mm}
    \caption{Controlled diagnostics. (a) Hard content projection also removes supported style evidence. (b) The effective residual budget governs when purified evidence becomes visible and when reference structure begins to dominate.}
    \label{fig:controlled}
\end{figure}

To this end, we propose SafeStyle, a training free framework that formulates reference stylization as safe evidence extraction followed by adaptive evidence transport. Contrastive calibration estimates style supported and content associated subspaces, preserves their useful overlap, and suppresses only unsupported content variations. The purified evidence is then transported through a unified operator over fine, middle, and coarse spatial supports according to style characteristics. Finally, an actual residual budget constrains the modification introduced into the frozen denoising network, preventing excessive reference influence. SafeStyle requires neither parameter updates nor explicit style words and uses a single generation path for both texture and geometry styles.
Our contributions are summarized as follows:
\begin{itemize}[topsep=2pt plus 1pt,itemsep=0pt,parsep=0pt,partopsep=0pt]
    \item We formulate reference stylization as safe evidence extraction and adaptive granularity-aware transport  in a frozen diffusion model, jointly addressing content leakage and style-dependent spatial organization.

    \item We identify persistent content-style subspace overlap and style-dependent spatial support, motivating protected purification and unified multi-granularity evidence transport.

    \item Experiments across texture and geometry styles show improved style fidelity and competitive text alignment, with 0.8\% leakage on a semantically disjoint stress benchmark.
\end{itemize}
\endgroup

\begin{figure*}[t!]
    \centering
    \includegraphics[width=0.965\linewidth]{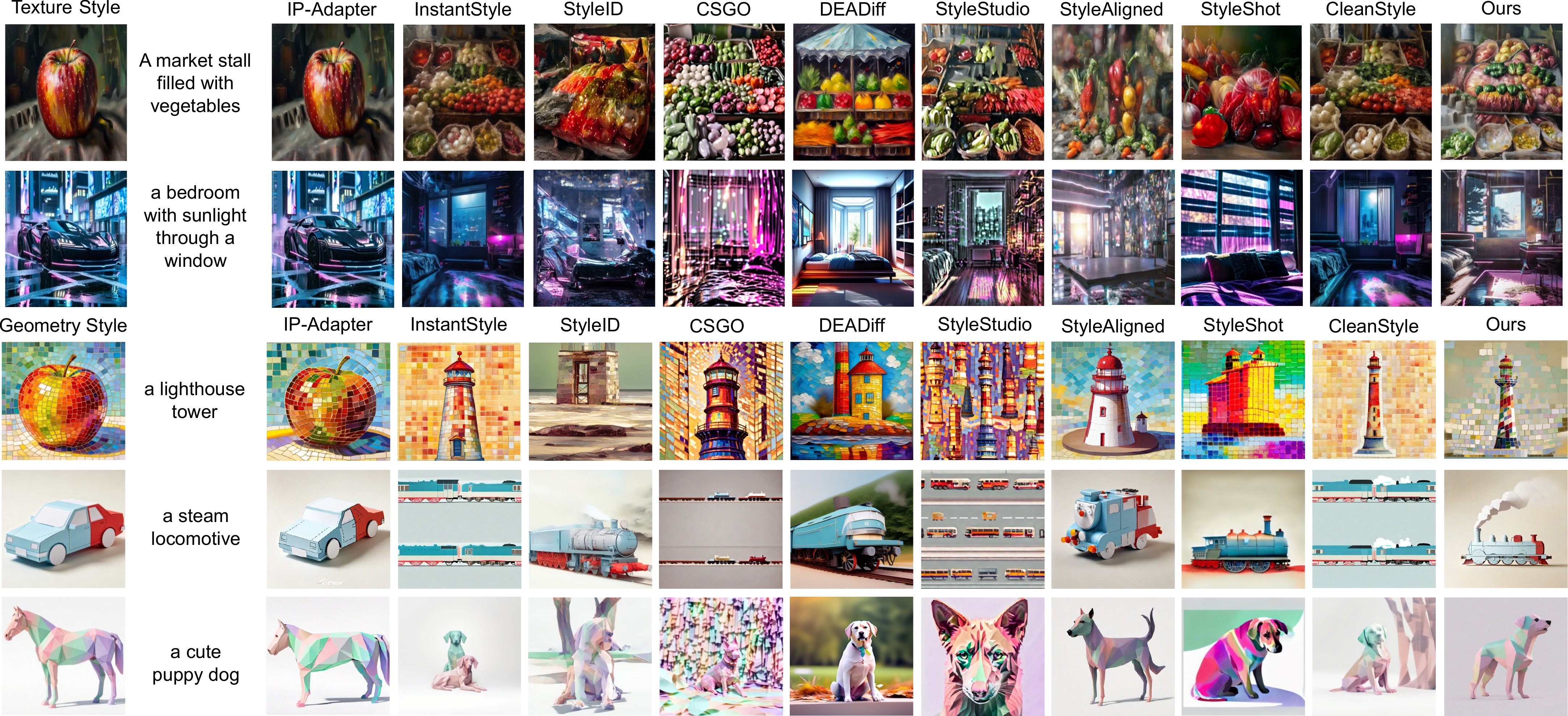}
    \vspace{-0.2cm}
    \caption{Representative texture (top) and geometry (bottom) comparisons. SafeStyle preserves the target subject while retaining stronger local appearance and structured style evidence.}
    \label{fig:qualitative}
        \vspace{-0.2cm}
\end{figure*}

\section{Method}
\label{sec:method}

\subsection{Overview}
\label{sec:overview}

SafeStyle is motivated by two observations about reference guided stylization. First, content and style are not orthogonally separable. As shown in Fig.~\ref{fig:diagnosis}(a), style associated and content associated subspaces exhibit a consistent nonzero overlap because visual cues in images can simultaneously describe object content and style. Fig.~\ref{fig:controlled}(a) further shows that directly removing the entire content subspace suppresses reference leakage but also discards useful strokes and structural style cues. Second, transferable style evidence exhibits different spatial organization across style types. As shown in Fig.~\ref{fig:diagnosis}(b), geometry dominant styles are more sensitive to spatial permutation, whereas texture dominant styles retain more local appearance information. A fixed fine scale transfer may weaken structured styles, while excessive coarse scale transfer may lead to leakage again. Fig.~\ref{fig:controlled}(b) additionally shows that proper budget injection improves stylization with high fidelity.

These observations suggest that reference stylization requires three coordinated decisions: which evidence should be preserved, at what spatial support it should be transported, and how strongly it should affect denoising. Based on this principle, SafeStyle receives a runtime reference $I_{\mathrm{ref}}$, a text prompt $p$, and compact offline style, content, and base calibration sets $\mathcal D_s$, $\mathcal D_c$, and $\mathcal D_b$. These calibration sets are constructed once and reused across references and prompts. A frozen image encoder first extracts reference tokens $E_{\mathrm{ref}}$. As illustrated in Fig.~\ref{fig:overview}, SafeStyle then performs contrastive style purification to preserve style supported overlap while suppressing unsupported content evidence. The surviving evidence is subsequently transported over adaptive spatial granularity, and its actual contribution to the frozen U-Net is constrained by a residual budget. In this way, evidence selection, spatial support, and injection strength are realized by three dedicated mechanisms, which nevertheless act jointly within a single frozen generation path.

\subsection{Contrastive Style Purification}
The style set contains different subjects sharing a style, while the content set contains the reference subject under varied appearances. After centering their token features, truncated SVD estimates the style- and content-associated bases $U_s$ and $U_c$:
\begin{equation}
\begin{aligned}
R_s &= E_s - \mu_b \simeq U_s \Sigma_s V_s^\top, \\
R_c &= E_c - \mu_c \simeq U_c \Sigma_c V_c^\top, \\
P_{c\setminus s} &= \bar U_c \bar U_c^\top,
\end{aligned}
\label{eq:bases}
\end{equation}
where $\bar U_c = \operatorname{orth}((I - U_s U_s^\top)U_c)$ is the orthonormal basis of the content subspace after projecting out the style components. Consequently, $P_{c\setminus s}$ captures only the content variations that lie strictly outside the style-supported subspace. For the centered reference token $v = e_{\mathrm{ref}} - \mu_s$, the purified token is computed as:
\begin{equation}
e_{\mathrm{clean}} = \mu_s + \beta U_s U_s^\top v + (I - \alpha P_{c\setminus s})(I - U_s U_s^\top)v,
\label{eq:purify}
\end{equation}
where the adaptive attenuation factor is $\alpha = \alpha_0(1-o)$, with the basis overlap ratio defined as $o = \lVert U_s^\top U_c\rVert_F^2 / \min(r_s, r_c)$. The first residual term preserves style-supported variation, while the second attenuates unsupported content. A greater basis overlap naturally decreases $\alpha$, preventing hard projection from discarding evidence shared by both style and content.

\subsection{Granularity-Aware Transport under a Residual Budget}
Purification decides which evidence survives; transport determines its spatial support and effective strength. SafeStyle uses the purified reference for fine appearance cues and compresses the style set into a prototype memory $\mathcal M_s=\{(k_i,v_i,\pi_i^s)\}_{i=1}^K$. The reference reweights safe prototypes through $\pi_i^\star\propto(\pi_i^s)^{1-\rho^\star}(\pi_i^{\mathrm{ref}})^{\rho^\star}$, where $\rho^\star$ is the largest coefficient permitted by a distribution budget. This retains reference-dependent palette and primitive preferences without replacing shared style statistics with reference layout.

All evidence is transported by the same operator:
\begin{equation}
r_{\ell,t}=\sum_{q\in\{f,m,c\}}g_q\,
\mathcal T_{\ell,t}^{q}(E_{\mathrm{clean}},\mathcal M_s^\star),
\label{eq:transport}
\end{equation}
where fine support conveys local color and strokes, middle support conveys repeated primitives or regions, and coarse support regulates low-frequency foreground-background organization. The profile $\mathbf g=(g_f,g_m,g_c)$ changes only transport resolution. When coarse evidence is needed, early denoising preserves target-related low frequencies inside the foreground role mask and gradually releases this protection after identity and object count stabilize.
Reference scale alone does not bound the change made to a U-Net feature. We therefore constrain the residual after all supports are aggregated:
\begin{equation}
\begin{aligned}
\widetilde r_{\ell,t}&=\min\!\left(1,
\frac{\eta\lVert h_{\ell,t}\rVert_2}
{\lVert r_{\ell,t}\rVert_2+\epsilon}\right)r_{\ell,t},\\
h_{\ell,t}^{\mathrm{out}}&=h_{\ell,t}
+\lambda_\ell\gamma_t\widetilde r_{\ell,t}.
\end{aligned}
\label{eq:budget}
\end{equation}
Thus, $\mathbf g$ controls transport granularity, while the budget limits the final relative feature change by $\eta\lambda_\ell\gamma_t$, keeping all three decisions within the same generation path.

\vspace{-2mm}

\section{Experiments}
\noindent\textbf{Setup.}
SafeStyle uses frozen SDXL with the InstantStyle adapter. We evaluate a curated StyleAdapter subset~\cite{StyleAdapter} containing 50 style references and 20 target prompts. Nine baselines are tested with identical references, prompts, and seeds: IP-Adapter, InstantStyle, StyleID, CSGO, DEADiff, StyleStudio, StyleAligned, StyleShot, and CleanStyle~\cite{CleanStyle}. We report CLIP text alignment (CLIP-TA), CLIP style similarity (CLIP-SS)~\cite{CLIP}, and DINO style similarity (DINO-SS)~\cite{DINO}. 
We observe that semantic overlap between reference and target subjects can implicitly mask content leakage. To further verify the leakage, we design a leakage stress set, which pairs 24 references with ten semantically disjoint prompts. Let $q(c)$ and $q(t)$ denote the CLIP text embeddings of the reference and target subjects, and $z_I(\hat{x})$ the CLIP image embedding of the generated output. We define
\begin{equation}
    \Delta_{\text{sem}} = \cos\big(z_I(\hat{x}), q(c)\big) - \cos\big(z_I(\hat{x}), q(t)\big).
\end{equation}
We evaluate leakage rate via the semantic leakage score (Sem-Leak). A sample is viewed as leaked when $\Delta_{\text{sem}} > 0$, indicating higher semantic alignment with the reference than the prompt. All experiments run on one RTX 4090.

\subsection{Main Comparisons}

\begin{table}[t]
\centering
\caption{Overall and leakage stress set comparisons. For brevity, C-SS, D-SS, and Leak denote CLIP-SS, DINO-SS, and Sem-Leak. Best results are in \textbf{bold}}
\label{tab:main}
\scriptsize
\setlength{\tabcolsep}{1.25pt}
\renewcommand{\arraystretch}{0.92}
\begin{tabular}{@{}lccc@{\hspace{2pt}}ccc@{}}
\toprule
&\multicolumn{3}{c}{Overall}&\multicolumn{3}{c}{Leakage Stress Set}\\
\cmidrule(lr){2-4}\cmidrule(l){5-7}
Method & TA$\uparrow$ & C-SS$\uparrow$ & D-SS$\uparrow$ & C-SS$\uparrow$ & D-SS$\uparrow$ & Leak$\downarrow$\\
\midrule
IP-Adapter~\cite{IPAdapter}       &0.085&0.753&0.454& 0.754&0.518&1.000\\
InstantStyle~\cite{InstantStyle}  &0.229&0.726&0.354& 0.711&0.361&0.364\\
StyleID~\cite{StyleID}            &0.221&0.709&0.320& 0.684&0.276&0.000\\
CSGO~\cite{CSGO}                  &0.233&0.701&0.304& 0.681&0.357&0.012\\
DEADiff~\cite{DEADiff}            &\textbf{0.243}&0.702&0.336& 0.684&0.384&0.004\\
StyleStudio~\cite{StyleStudio}    &0.241&0.691&0.290& 0.671&0.305&0.000\\
StyleAligned~\cite{StyleAligned}  &0.185&0.716&0.377& 0.712&0.399&0.258\\
StyleShot~\cite{StyleShot}        &0.222&0.714&0.345& 0.703&0.356&0.008\\
CleanStyle~\cite{CleanStyle}      &0.235&0.723&0.342& 0.699&0.349&0.000\\
\textbf{SafeStyle (Ours)}         &0.236&\textbf{0.729}&\textbf{0.432}& \textbf{0.720}&\textbf{0.474}&0.008\\
\bottomrule
\end{tabular}
\end{table}
\begin{table}[t]
\centering
\caption{Component ablation on the leakage-stress benchmark.}
\label{tab:ablation}
\scriptsize
\setlength{\tabcolsep}{2.4pt}
\renewcommand{\arraystretch}{0.92}
\begin{tabular}{@{}lccc@{}}
\toprule
Configuration & CLIP-TA$\uparrow$ & DINO-SS$\uparrow$ & Sem-Leak$\downarrow$\\
\midrule
Raw evidence       &0.190&0.485&0.279\\
Hard projection    &0.224&0.445&0.008\\
Safe purification  &0.198&\textbf{0.530}&0.167\\
$+$ Transport      &0.235&0.498&0.029\\
$+$ Budget (Ours)  &\textbf{0.238}&0.474&\textbf{0.008}\\
\bottomrule
\end{tabular}
\vspace{0.5mm}
\end{table}

\noindent\textbf{Qualitative Results.} Fig.~\ref{fig:qualitative} compares baselines across texture- and geometry-dominant styles. IP-Adapter~\cite{IPAdapter} is prone to reference leakage, indicating strong content-style coupling. InstantStyle~\cite{InstantStyle} and CleanStyle~\cite{CleanStyle} better preserve the target semantics but transfer brushwork and geometric primitives conservatively. StyleID~\cite{StyleID} and DEADiff~\cite{DEADiff} weaken stylization or revert structured styles toward realistic objects. CSGO~\cite{CSGO} and StyleShot~\cite{StyleShot} retain salient colors and local patterns, yet their geometric organization remains inconsistent. StyleStudio~\cite{StyleStudio} produces dense or repeated structures, while StyleAligned~\cite{StyleAligned} occasionally preserves reference-specific shapes, such as the car-like locomotive. SafeStyle consistently follows the target prompts while retaining visual effects of styles, achieving the best balance between style fidelity and reference-content suppression across both domains.

\noindent\textbf{Quantitative Results.} The overall scores in Tab~\ref{tab:main} support the visual observations. DEADiff~\cite{DEADiff} achieves the highest CLIP-TA score(0.243) at the cost of style transfer failures, while SafeStyle remains competitive at 0.236. SafeStyle reaches 0.729 CLIP-SS and 0.432 DINO-SS, outperforming all baselines in style similarity except IP-Adapter~\cite{IPAdapter}, whose higher CLIP-SS and DINO-SS coincide with a collapsed CLIP-TA of 0.085. The leakage-stress results in Sec.~3.4 can confirm that this apparent advantage in style similarity stems from reference copying rather than transferable style evidence. Compared with CleanStyle~\cite{CleanStyle}, it improves CLIP-SS and DINO-SS by 0.006 and 0.090 without reducing text alignment.

\begin{figure}[t!]
    \centering
    \includegraphics[width=0.96\linewidth]{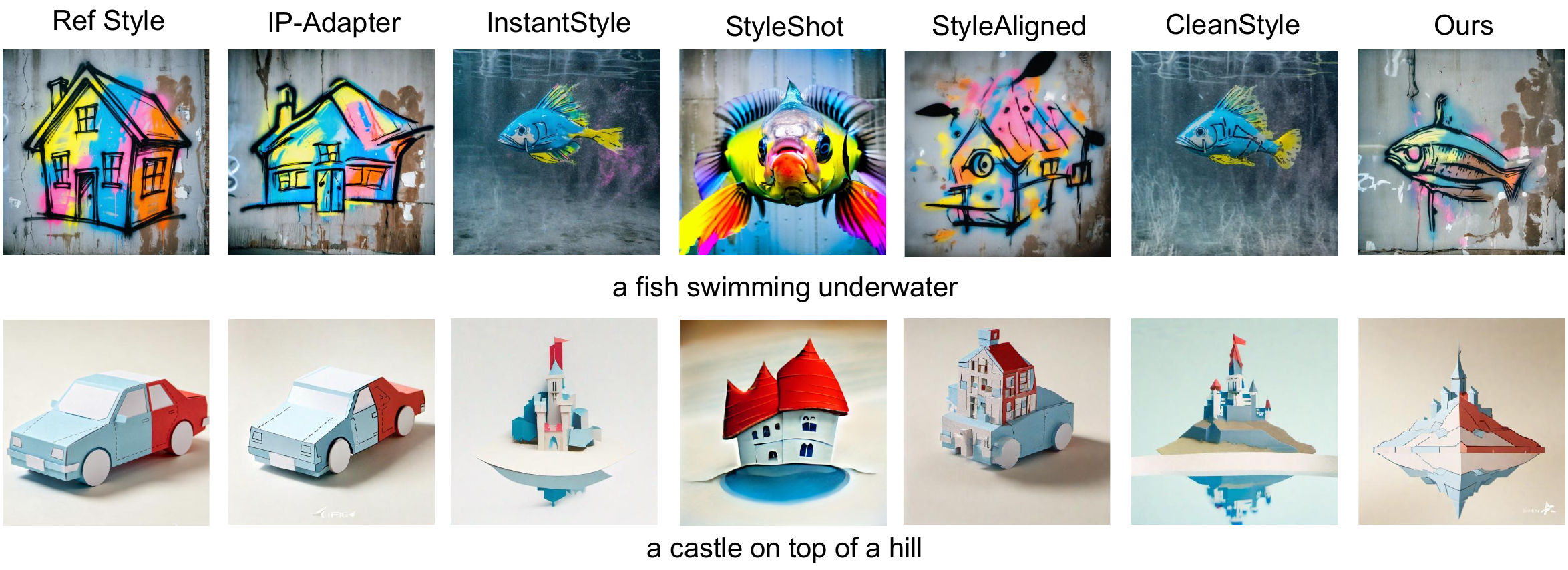}
    \vspace{-4mm}
    \caption{Leakage-stress examples with semantically disjoint reference and target subjects. SafeStyle retains stronger style evidence while suppressing subject transfer.}
    \label{fig:stress}
    \vspace{-2mm}
\end{figure}
\begin{figure}[t!]
    \centering
    \includegraphics[width=0.96\linewidth]{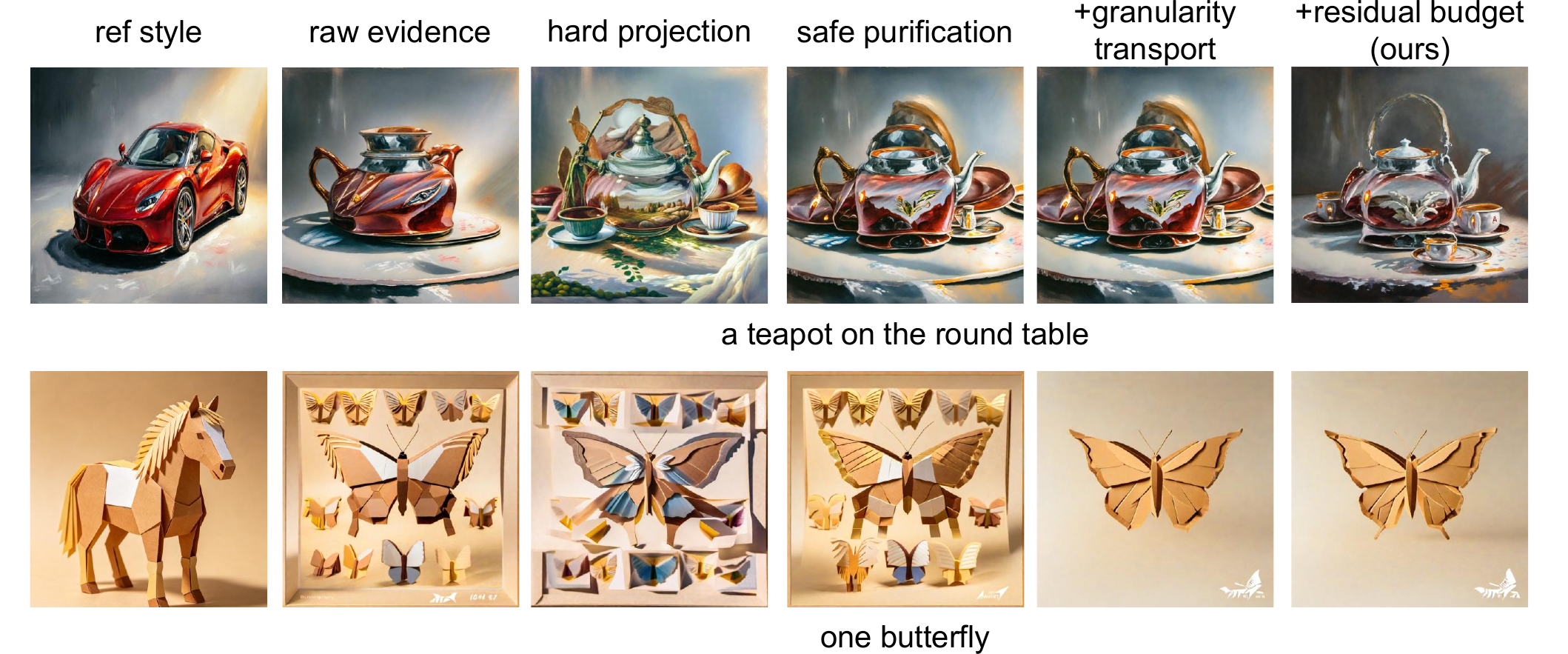}
    \vspace{-3mm}
    \caption{Unified ablation on oil-painting and paper craft references. The same purification, granularity transport, and residual budget are evaluated across both style groups.}
    \label{fig:ablation}

\end{figure}

\subsection{Ablation Study}
We ablate SafeStyle by adding modules stepwise. As shown in Tab~\ref{tab:ablation}, raw evidence retains high style similarity with unacceptable leakage(0.279). Hard projection reduces leakage to 0.008, while its DINO-SS decreases to 0.445 because transferable style evidence is also removed. Safe purification preserves the shared style-content support and raises DINO-SS to 0.530, but the remaining 0.167 leakage shows that purification alone cannot regulate its influence. Granularity-aware transport significantly increases CLIP-TA(+0.037) and reduces leakage(-0.138). The residual budget further lowers leakage to 0.008 while retaining 0.474 DINO-SS. Fig.~\ref{fig:ablation} shows that transport restores the single-butterfly composition, while budgeting removes the car-like teapot structure without discarding the paper folds or oil-painting appearance.

\noindent\textbf{Leakage-Stress Results.} As illustrated in Fig.~\ref{fig:stress}, raw evidence injection leads to the leakage, while aggressive filtering weakens the texture strokes or geometric construction. SafeStyle instead preserves fish and castle while retaining the graffiti marks and paper craft structure. On the leakage stress set of Tab~\ref{tab:main}, SafeStyle achieves 0.720 CLIP-SS and 0.474 DINO-SS with a leakage rate of 0.008. At the same leakage rate as StyleShot~\cite{StyleShot}, it improves the two style metrics by 0.017 and 0.118. CleanStyle~\cite{CleanStyle} eliminates detected leakage but reduces style similarity largely. SafeStyle therefore trades only 0.8\% semantic leakage for gains of 0.021 in CLIP-SS and 0.125 in DINO-SS, providing a best style-content balance.
\vspace{-2.5mm}

\section{Conclusion}
We presented SafeStyle, a training-free framework that formulates reference stylization as safe evidence extraction and granularity-aware residual injection under an explicit residual budget. Protecting the style-supported content overlap, transporting purified evidence at adaptive spatial supports, and bounding its actual residual together yield the best style fidelity among methods without systematic reference copying, with competitive text alignment.

\balance
\bibliographystyle{IEEEbib}
\bibliography{strings,references}

\begin{thebibliography}{10}

\bibitem{LDM}
Robin Rombach, Andreas Blattmann, Dominik Lorenz, Patrick Esser, and Bj{\"o}rn Ommer,
\newblock ``High-resolution image synthesis with latent diffusion models,''
\newblock in {\em Proceedings of the IEEE/CVF Conference on Computer Vision and Pattern Recognition}, 2022, pp. 10684--10695.

\bibitem{SDXL}
Dustin Podell, Zion English, Kyle Lacey, Andreas Blattmann, Tim Dockhorn, Jonas M{\"u}ller, Joe Penna, and Robin Rombach,
\newblock ``{SDXL}: Improving latent diffusion models for high-resolution image synthesis,''
\newblock in {\em International Conference on Learning Representations}, 2024.

\bibitem{IPAdapter}
Hu~Ye, Jun Zhang, Sibo Liu, Xiao Han, and Wei Yang,
\newblock ``{IP-Adapter}: Text compatible image prompt adapter for text-to-image diffusion models,''
\newblock {\em arXiv preprint arXiv:2308.06721}, 2023.

\bibitem{InstantStyle}
Haofan Wang, Matteo Spinelli, Qixun Wang, Xu~Bai, Zekui Qin, and Anthony Chen,
\newblock ``{InstantStyle}: Free lunch towards style-preserving in text-to-image generation,''
\newblock {\em arXiv preprint arXiv:2404.02733}, 2024.

\bibitem{DEADiff}
Tianhao Qi, Shancheng Fang, Yanze Wu, Hongtao Xie, Jiawei Liu, Lang Chen, Qian He, and Yongdong Zhang,
\newblock ``{DEADiff}: An efficient stylization diffusion model with disentangled representations,''
\newblock in {\em Proceedings of the IEEE/CVF Conference on Computer Vision and Pattern Recognition}, 2024, pp. 8693--8702.

\bibitem{CSGO}
Peng Xing, Haofan Wang, Yanpeng Sun, Qixun Wang, Xu~Bai, Hao Ai, Renyuan Huang, and Zechao Li,
\newblock ``{CSGO}: Content-style composition in text-to-image generation,''
\newblock in {\em Advances in Neural Information Processing Systems}, 2025, vol.~38, pp. 111464--111504.

\bibitem{StyleStudio}
Mingkun Lei, Xue Song, Beier Zhu, Hao Wang, and Chi Zhang,
\newblock ``{StyleStudio}: Text-driven style transfer with selective control of style elements,''
\newblock in {\em Proceedings of the IEEE/CVF Conference on Computer Vision and Pattern Recognition}, 2025, pp. 23443--23452.

\bibitem{StyleShot}
Junyao Gao, Yanan Sun, Yanchen Liu, Yinhao Tang, Yanhong Zeng, Ding Qi, Kai Chen, and Cairong Zhao,
\newblock ``{StyleShot}: A snapshot on any style,''
\newblock {\em IEEE Transactions on Pattern Analysis and Machine Intelligence}, vol. 48, no. 2, pp. 1215--1228, 2026.

\bibitem{StyleID}
Jiwoo Chung, Sangeek Hyun, and Jae-Pil Heo,
\newblock ``Style injection in diffusion: A training-free approach for adapting large-scale diffusion models for style transfer,''
\newblock in {\em Proceedings of the IEEE/CVF Conference on Computer Vision and Pattern Recognition}, 2024, pp. 8795--8805.

\bibitem{StyleAligned}
Amir Hertz, Andrey Voynov, Shlomi Fruchter, and Daniel Cohen-Or,
\newblock ``Style aligned image generation via shared attention,''
\newblock in {\em Proceedings of the IEEE/CVF Conference on Computer Vision and Pattern Recognition}, 2024, pp. 4775--4785.

\bibitem{MaskST}
Lin Zhu, Xinbing Wang, Chenghu Zhou, Qinying Gu, and Nanyang Ye,
\newblock ``Less is more: Masking elements in image condition features avoids content leakages in style transfer diffusion models,''
\newblock in {\em International Conference on Learning Representations}, 2025.

\bibitem{StyleKeeper}
Jaeseok Jeong, Junho Kim, Gayoung Lee, Yunjey Choi, and Youngjung Uh,
\newblock ``{StyleKeeper}: Prevent content leakage using negative visual query guidance,''
\newblock in {\em Proceedings of the IEEE/CVF International Conference on Computer Vision}, 2025, pp. 15760--15769.

\bibitem{StyleSSP}
Ruojun Xu, Weijie Xi, XiaoDi Wang, Yongbo Mao, and Zach Cheng,
\newblock ``{StyleSSP}: Sampling startpoint enhancement for training-free diffusion-based method for style transfer,''
\newblock in {\em Proceedings of the IEEE/CVF Conference on Computer Vision and Pattern Recognition}, 2025, pp. 18260--18269.

\bibitem{OSASIS}
Hansam Cho, Jonghyun Lee, Seunggyu Chang, and Yonghyun Jeong,
\newblock ``One-shot structure-aware stylized image synthesis,''
\newblock in {\em Proceedings of the IEEE/CVF Conference on Computer Vision and Pattern Recognition}, 2024, pp. 8302--8311.

\bibitem{imagdressing}
Fei Shen, Xin Jiang, Xin He, Hu~Ye, Cong Wang, Xiaoyu Du, Zechao Li, and Jinhui Tang,
\newblock ``Imagdressing-v1: Customizable virtual dressing,''
\newblock in {\em Proceedings of the AAAI Conference on Artificial Intelligence}, 2025, vol.~39, pp. 6795--6804.

\bibitem{imagpose}
Fei Shen and Jinhui Tang,
\newblock ``Imagpose: A unified conditional framework for pose-guided person generation,''
\newblock {\em Advances in neural information processing systems}, vol. 37, pp. 6246--6266, 2024.

\bibitem{StyleGallery}
Boyu He, Yunfan Ye, Chang Liu, Weishang Wu, Fang Liu, and Zhiping Cai,
\newblock ``{StyleGallery}: Training-free and semantic-aware personalized style transfer from arbitrary image references,''
\newblock in {\em Proceedings of the IEEE/CVF Conference on Computer Vision and Pattern Recognition}, 2026, pp. 29092--29102.

\bibitem{SEFS}
Jingtao Zhang, Haorui Gao, Youqing Liang, and Zeming Liu,
\newblock ``Scale-separated conditioning for style-encoder-free diffusion stylization,''
\newblock {\em arXiv preprint arXiv:2608.19719}, 2026.

\bibitem{CleanStyle}
Xiaoman Feng, Mingkun Lei, Yang Wang, Dingwen Fu, and Chi Zhang,
\newblock ``{CleanStyle}: Plug-and-play style conditioning purification for text-to-image stylization,''
\newblock {\em arXiv preprint arXiv:2602.20721}, 2026.

\bibitem{StyleAdapter}
Zhouxia Wang, Xintao Wang, Liangbin Xie, Zhongang Qi, Ying Shan, Wenping Wang, and Ping Luo,
\newblock ``{StyleAdapter}: A unified stylized image generation model,''
\newblock {\em International Journal of Computer Vision}, vol. 133, no. 4, pp. 1894--1911, 2025.

\bibitem{CLIP}
Alec Radford, Jong~Wook Kim, Chris Hallacy, Aditya Ramesh, Gabriel Goh, Sandhini Agarwal, Girish Sastry, Amanda Askell, Pamela Mishkin, Jack Clark, Gretchen Krueger, and Ilya Sutskever,
\newblock ``Learning transferable visual models from natural language supervision,''
\newblock in {\em Proceedings of the 38th International Conference on Machine Learning}, 2021, vol. 139, pp. 8748--8763.

\bibitem{DINO}
Mathilde Caron, Hugo Touvron, Ishan Misra, Herv{\'e} J{\'e}gou, Julien Mairal, Piotr Bojanowski, and Armand Joulin,
\newblock ``Emerging properties in self-supervised vision transformers,''
\newblock in {\em Proceedings of the IEEE/CVF International Conference on Computer Vision}, 2021, pp. 9650--9660.

\end{thebibliography}

\end{document}